\documentclass[12pt]{article}

\RequirePackage{kpfonts}
\RequirePackage[sf,bf,small,raggedright]{titlesec}
\usepackage{dsfont}
\usepackage{comment}
\usepackage{textcomp}
\usepackage{amssymb}
\usepackage{bbm}
\usepackage{fancyhdr}
\usepackage{amsmath}
\usepackage{amsbsy,amsthm}
\usepackage{amscd}
\usepackage{latexsym}
\usepackage{graphicx}   
\usepackage{pdfsync}
\usepackage{blkarray}
\usepackage{multirow}
\usepackage{hyperref}
\usepackage{xcolor}
\usepackage{enumerate}

\usepackage{natbib}
\usepackage{tcolorbox}

\usepackage{tikz}

\usepackage{subcaption}
\usepackage{algorithmic}
\usepackage[justification=centering]{caption}

\usepackage{natbib}
 \def\newblock{\ }%
 \bibpunct[, ]{[}{]}{,}{n}{}{,}%

\usepackage{float}
\usepackage{pgf}
\usetikzlibrary{arrows}

\newcommand{\R}{\mathbb{R}}

\newtheorem{theorem}{Theorem}
\newtheorem{lemma}[theorem]{Lemma}

\allowdisplaybreaks

\numberwithin{equation}{section}

\def \proof {\noindent {\bf Proof.}\ \ }

\def\IND{\mathbbm{1}}

\newcommand{\argmin}{\mathop{\mathrm{argmin}}}

\newcommand{\EXP}{\mathbb{E}}
\newcommand{\PROB}{\mathbb{P}}

\newcommand{\Var}{\mathrm{Var}}

\newcommand{\defeq}{\stackrel{\mathrm{def.}}{=}}
\newcommand{\inlaw}{\stackrel{\mathcal{L}}{=}}

\begin{document}

\title{A note on estimating the dimension from a random geometric graph
  \thanks{
Luc Devroye acknowledges support of NSERC grant A3450.   
  G\'abor Lugosi acknowledges the support of Ayudas Fundación BBVA a
  Proyectos de Investigación Científica 2021 and
of the Spanish grant PID2022-138268NB-I00, financed by
MCIN/AEI/10.13039/501100011033, FSE+MTM2015-67304-P, and FEDER, EU.
}
\author{
Caelan Atamanchuk\\
   School of Computer Science\\
  McGill University \\
  Montreal, Canada
  \and
  Luc Devroye  \\
  School of Computer Science\\
  McGill University \\
  Montreal, Canada
\and  
G\'abor Lugosi \\
Department of Economics and Business, \\
Pompeu  Fabra University, Barcelona, Spain \\
ICREA, Pg. Lluís Companys 23, 08010 Barcelona, Spain \\
Barcelona Graduate School of Economics
}
}

\maketitle

\begin{abstract}
Let $G_n$ be a  random geometric graph with vertex set $[n]$
based on $n$ i.i.d.\ random vectors $X_1,\ldots,X_n$ drawn
from an unknown density $f$ on $\R^d$. An edge $(i,j)$ is present when
$\|X_i -X_j\| \le r_n$, for a given threshold $r_n$ possibly depending upon $n$,
where $\| \cdot \|$ denotes Euclidean distance.
We study the problem of estimating the dimension $d$ of the underlying
space when we have access to the adjacency matrix of the graph but
 do not know $r_n$ or the vectors $X_i$.
The main result of the paper is that
there exists an estimator of $d$ that converges
to $d$ in probability as $n \to \infty$ for all densities with $\int f^5 < \infty$
whenever $n^{3/2} r_n^d \to \infty$ and $r_n = o(1)$.
The conditions allow very sparse graphs since when $n^{3/2} r_n^d \to 0$, the graph contains
isolated edges only, with high probability.
We also show that, without any condition on the density, a consistent estimator of $d$
exists 
when $n r_n^d \to \infty$ and $r_n = o(1)$.
\end{abstract}


\section{Introduction}

In network science one often seeks geometric representations of an observed network
that help interpret and predict connections and understand the structure of the network.
Indeed, significant effort has been devoted to embedding vertices of a graph in Euclidean
or hyperbolic spaces, see
Reiterman, R{\"o}dl, {\v{S}}i{\v{n}}ajov{\'a} \cite{ReRoSi89},
Tenenbaum, Silva, and Langford \cite{TeSiLa00},
Shavitt and Tankel \cite{ShTa04}, Kleinberg \cite{Kle07}, Kang and M\"uller \cite{KaMu11}, Verbeek and Suri \cite{VeSu14}
for a sample of the literature. A more basic question is to determine the dimension
of the underlying geometric space. 
In this paper we consider the problem of estimating the
dimension of the Euclidean space underlying a geometric graph, upon observing a (combinatorial)
graph.

In order to set up a rigorous statistical problem, we model the graph as a random geometric
graph.
Let $G_n$ be a  random geometric graph with vertex set $[n]$
based on $n$ i.i.d.\ random vectors $X_1,\ldots,X_n$ drawn
from an unknown density $f$ on $\R^d$. An edge $(i,j)$ is present when
$\|X_i -X_j\| \le r_n$, for a given threshold $r_n$ possibly depending upon $n$,
where $\| \cdot \|$ denotes Euclidean distance.
Introduced by Gilbert \cite{Gil59,Gil61}, the properties of these graphs
have been well studied when $f$ is the uniform density on a convex set of $\R^d$
or the torus $[0,1]^d$ in $\R^d$. Its properties are surveyed
by Penrose \cite{Pen03}.
Noteworthy are the precise results on
connectivity (Appel and Russo \cite{ApRu02};
Balister, Bollob\'as and Sarkar \cite{BaBoSa08};
Balister, Bollob\'as, Sarkar, and Walters \cite{BaBoSaWa09}),
cover time (Cooper and Frieze \cite{CoFr11}),
coverage (Gilbert \cite{Gil65}; Hall \cite{Hal88}, Janson \cite{Jan86}),
chromatic number (McDiarmid and M\"uller \cite{McMu11}),
and minimal spanning tree (Penrose \cite{Pen99}).

In the dimension-estimation problem considered here,
we observe the adjacency matrix of the graph $G_n$
but we do not know $d$, $r_n$ or the vertex locations $X_i$.
The question then is whether one can estimate the underlying dimension $d$.
In other words, can one develop an estimate $\Delta_n$ of $d$, based only
on knowledge of $G_n$, with the property that $\Delta_n \to d$ in probability 
as $n \to \infty$? When this convergence happens, we say that $\Delta_n$ is a
\emph{consistent} estimator of $d$.

Whether consistent estimators exist, may depend on the parameters of the model, that is,
the density $f$ and the sequence of radii $\{r_n\}$. For example, if the graph is too sparse,
there is no hope to estimate $d$. Indeed, suppose that $f$ is the uniform density on $[0,1]^d$, 
and $r_n$ is such that $n^{3/2} r_n^d \to 0$. Then $G_n$ only contains isolated edges, with high
probability. Indeed,
\begin{eqnarray*}
\lefteqn{  
  \PROB\left\{ \exists \ \text{distinct} \  i,j,k \in [n]: \|X_i -X_j\| \le r_n \ \text{and} \ \|X_i -X_k\| \le r_n \right\}  } \\
&  \le &
  n^3 \EXP\left[ \PROB\left\{ \|X_1 -X_2\| \le r_n \ \text{and} \ \|X_1 -X_3\| \le r_n \Big{|} X_1 \right\}\right] 
 \le  n^3V_d^2 r_d^{2d} \to 0~,
\end{eqnarray*}
where $V_d$ denotes the volume of the unit ball in $\R^d$

For such graphs it is clearly impossible to infer anything about the underlying geometry. 
The main result of this paper shows that, as soon as $r_n$ is such that $n^{3/2} r_n^d \to \infty$,
it is possible to consistently estimate the dimension, for a large class of densities.
More precisely, we prove the following.

\begin{theorem}
  \label{thm1}
  Let the density $f$ on $\R^d$ satisfy $\int f^5 < \infty$. Assume furthermore that
$$
\lim_{n \to \infty} n^{3/2} r_n^d = \infty~,
$$
and $r_n = o(1)$.
Then there exists an estimate $\Delta_n$ such that $\Delta_n \to d$ in probability.
\end{theorem}

The condition on the radius $r_n$ allows extremely sparse graphs. It suffices to have $r_n^d \sim n^{-3/2}\omega_n$ for $\omega_n \to \infty$ arbitrarily slowly. Note that in that case
the graph has merely $O_p(\sqrt{n}\omega_n)$ edges.
The condition $\int f^5 < \infty$ excludes densities with pronounced infinite peaks but
it does not assume anything about the smoothness or tails of the distribution.
We also prove a consistency result for arbitrary densities, though under more stringent
conditions on the radii $r_n$:

\begin{theorem}
\label{thm2}
Let the density $f$ in $\R^d$ be arbitrary, and assume that
$$
\lim_{n \to \infty} n r_n^d = \infty~, 
\quad \text{and} \quad \lim_{n \to \infty} r_n = 0~.
$$
Then there exists an estimate $\Delta_n$ such that $\Delta_n \to d$ in probability.
\end{theorem}

It is an interesting open question whether there exist dimension estimators
that are consistent for all densities under the minimal assumption $n^{3/2}r_n^d \to \infty$.
We conjecture that the estimator used in this paper to prove Theorem \ref{thm1}
is not consistent for all densities, though the condition $\int f^5 < \infty$ may possibly be relaxed to
$\int f^3 < \infty$, as discussed below.

The paper is organized as follows. After reviewing some of the related literature,
in Section \ref{sec:geo} we establish a geometric lemma that is a key tool in our
approach of defining estimators of the dimension.
In Section \ref{sec:est} we introduce four simple estimators of the dimension whose analysis
proves Theorems \ref{thm1} and \ref{thm2}.
We start analyzing the estimators in Section \ref{sec:unif} by focusing on the special--but important--case
of the uniform density on the torus.
Finally, in Section \ref{sec:gen} we prove Theorems \ref{thm1} and \ref{thm2} for general densities.

\subsection*{Related literature}

Granata and Carnevale \cite{GrCa16} consider the dimension-estimation problem in a
more general framework of estimating the intrinsic dimension of geometric graphs defined in
general metric spaces. Instead of focusing on general conditions for consistency, \cite{GrCa16}
aim to construct accurate estimates from graph distances.

Bubeck, Ding, Eldan, and R{\'a}cz \cite{BuDiElRa16} show that, based on a dense random geometric graph
drawn from the uniform distribution on the surface of the $d$-dimensional unit sphere, 
it is possible to estimate $d$ as long as $n\gg d$.

Lichev, and Mitsche \cite{LiMi21} and Casse \cite{Cas23} study properties of the online nearest
neighbor tree based on uniformly distributed points in $[0,1]^d$
and observe that it is possible to consistently estimate the dimension upon observing the
combinatorial tree.

Dimension-estimation from geometric graphs is closely related to the problem of estimating
the intrinsic dimensionality of high-dimensional data. Indeed, often the first step of
computing such estimates is to construct a geometric graph from the data, see, e.g.,
Tenenbaum, Silva, and Langford \cite{TeSiLa00}, Facco, d’Errico, Rodriguez, Laio \cite{FaErRoLa17}.

Araya and De Castro \cite{ArDe19} study estimating the Euclidean distances between the
point locations upon observing the combinatorial graph for dense random geometric graphs.

\section{A geometric lemma}
\label{sec:geo}

Consider the unit ball $B(0,1)$ in $\R^d$, and let $X$ and $Y$ be independent and
uniformly distributed in $B(0,1)$. We define the quantity
$$
w_d = \PROB \{ \| X - Y \| \le 1 \}~.
$$
As all estimators studied in this paper are based on estimating $w_d$, 
the following property, proved in the Appendix, is a key ingredient of our arguments.

\begin{lemma}
 \label{lem:beta}
We have
$$
w_d 
= {3 \over 2} \PROB \left\{ \beta \left( {1 \over 2}, {d+1 \over 2} \right) \ge {1 \over 4} \right\}~,
$$
where $\beta (a,b)$ denotes a beta random variable with shape parameters $a$ and $b$.
The sequence $w_d$ decreases strictly monotonically to $0$ as $d \uparrow \infty$.
\end{lemma}

\subsubsection*{On the computation of $w_d$.}
The explicit density of $\|X-Y\|$ was derived by
Aharonyan and Khalatyan \cite{AhKh20}. From it, one can deduce a formula for $w_d$
as a function of some gamma functions.
As we showed in Lemma \ref{lem:beta}, the constant $w_d$ is simply related to the upper tail of a beta random variable,
so $w_d$ is a constant times an incomplete beta integral.
For general proporties of random variables uniformly distributed in high-dimensional convex sets,
we refer to Vershynin \cite{Ver18}.

\subsubsection*{On the sample size needed.}
The representation of $w_d$ given in Lemma \ref{lem:beta}
permits us to show that $w_d - w_{d+1} \ge d^{-(d+o(d))/2}$ (see the Appendix).
Our proposed algorithms are all based on estimates of $w_d$, and have errors
that decline at polynomial rates in $n$, the sample size.  Thus, while all estimates
are consistent in the limit, there is no hope of a good performance when $d \gg \log n / \log \log n$.
In \cite{BuDiElRa16} it is shown that, in the case of very dense graphs and the uniform density
on the surface of the unit sphere, there exist estimators that work well as soon as $n \gg d$.
It is a challenging problem for further research to determine the exact tradeoff between 
edge density and required sample size for accurately estimating $d$.

\section{The proposed estimates}
\label{sec:est}

Here we introduce four simple estimators of the quantity $w_d$ defined above. 
If $W$ is a data-based estimate, then we set
$$
\Delta_n = \argmin_d  | W - w_d | .
$$
In view of Lemma \ref{lem:beta}, if $W \to w_d$ in probability, then
$\Delta_n \to d$ in probability and therefore it suffices to construct consistent estimators of $w_d$.

By using binary search (first doubling the dimension until an overshoot occurs, and then applying
 classical binary search), one can find $\Delta_n$ using only $O(\log d)$ computations
of the function $w_s$.

We propose simple local estimates $W_1$ and $W_4$
and more powerful global estimates $W_2$ and $W_3$.
Randomly label the nodes of the graph
such that all labelings are equally likely.
Denote the degree of vertex $i$ in $G_n$ by $D_i$, and let
$\delta_i$ be the number of edges between nodes in $N_i$, the set of neighbors of vertex $i$.  
Let $M$ be the smallest index among the vertices of maximal degree.
Let $\xi_{ij}$ be the indicator that $i$ is connected to $j$.
Our estimates are as follows:
$$
W_1 \defeq {\delta_M \over {D_M \choose 2}}~,
$$
$$
W_2 \defeq {\sum_{i < j < k} \xi_{ki} \xi_{kj} \xi_{ij} \over \sum_{i < j < k} \xi_{ki} \xi_{kj} }~,
$$
$$
W_3 \defeq {\sum_{i=1}^n {\delta_i \over {D_i \choose 2}} \over \sum_{i=1}^n \IND_{D_i \ge 2} }~,
$$
and
$$
W_4 \defeq { \delta_1 \over {D_1 \choose 2} }~.
$$

\section{Analysis for the uniform density on the torus}
\label{sec:unif}

In this section we focus on the uniform density on $[0,1]^d$, 
and measure Euclidean distances
as in the torus, that is, for $x,y \in [0,1]^d$, 
$$
\| x - y \| \defeq \min_{z \in Z^d} \| x-y + z \|~,
$$
where $Z^d$ is the collection of all integer-valued $d$-dimensional vectors.
This allows us to present some of the ideas in a transparent manner.

Assume that $r_n \le 1/2$ to avoid the wraparound effect in the torus. 
We begin by analyzing the estimator $\delta_1$, assuming that $D_1 \ge 2$.
Note that we can represent $\delta_1$ as
$$
\delta_1 \inlaw \sum_{i<j \le D_1} Y_{ij}~,
$$
where $Y_{ij} = \IND_{ \| X_i - X_j \| \le 1}$,
and $X_1,\ldots,X_{D_1}$ are i.i.d.\ random vectors uniformly distributed in the unit ball of $\R^d$.
Each random variable $Y_{ij}$ is Bernoulli $(w_d)$.
Thus, still for $D_1 \ge 2$,
$$
\EXP \left\{ {\delta_1 \over {D_1 \choose 2} } \mid D_1  \right\} = w_d~,
$$
so that the estimator $\delta_1$ is unbiased. 
Then,
\begin{eqnarray*}
\Var \{ \delta_1 | D_1 \} 
&=& \EXP \left\{ \left( \sum_{i<j \le D_1} (Y_{ij}-w_d) \right)^2 \mid D_1 \right\} \\
&= & {D_1 \choose 2} \EXP \left\{ (Y_{12}-w_d)^2 | D_1\right\}
 +  3 {D_1 \choose 3}  \EXP \left\{ (Y_{12}-w_d) (Y_{13}-w_d) |D_1\right\}~. 
\end{eqnarray*}
Note however that, given $D_1$, $Y_{12}$ and $Y_{13}$ are conditionally independent, so that we can conclude that
$$
\Var \{ \delta_1 | D_1 \} 
= {D_1 \choose 2} w_d (1-w_d)~.
$$
Therefore,  by the Chebyshev-Cantelli inequality, if $D_1 \ge 2$ and $t > 0$,
\begin{eqnarray*}
\PROB \left\{ \left| {\delta_1 \over {D_1 \choose 2} } - w_d \right| \ge t \mid D_1 \right\}
&\le &  \Var \{ \delta_1 | D_1 \} \over \Var \{ \delta_1 | D_1 \} + {D_1 \choose 2}^2 t^2  \\
&= &  {D_1 \choose 2} w_d (1-w_d)  \over {D_1 \choose 2} w_d (1-w_d)  + {D_1 \choose 2}^2 t^2 \\
&= &  w_d (1-w_d)  \over w_d (1-w_d)  + {D_1 \choose 2} t^2  \\
&\le &  w_d  \over w_d + {D_1 \choose 2} t^2~. 
\end{eqnarray*}
We conclude that $W_4 = \delta_1/ {D_1 \choose 2} \to w_d$ in probability when $D_1 \to \infty$
in probability.

The following theorem summarizes the consistency properties of the estimators $W_1$ and $W_2$.

\begin{theorem}
  Let the density $f$ be the uniform density on the unit torus $[0,1]^d$, and assume that $r_n \le 1/2$ for all $n$.
\begin{itemize}
\item[(i)] If
$$
\lim_{n \to \infty} nr_n^{d(1-\epsilon)} = \infty
$$
for all $\epsilon > 0$, then $W_1 \to w_d$ (and thus $\Delta_n \to d$) in probability as $n \to \infty$. 
A sufficient condition for this is that $n r_n^d \ge L(n)$
with $L(n)$ slowly varying.
\item[(ii)] If $r_n \to 0$ and $n^{3/2} r_n^d \to \infty$, 
  then $W_2 \to w_d$ (and thus $\Delta_n \to d$) in probability as $n \to \infty$.
\end{itemize}
\end{theorem}

The computational complexity of the inferior estimate $W_1$ is less than that
of $W_2$, so both estimates have their use. On the other hand,
$W_1$ requires at least $n /L(n)$ edges, where $L(n)$  is slowly varying.
For example, for constant $k$, $n/ \log^k (n)$ edges will do.

\medskip
\noindent
{\bf Proof of (i).}
Replacing $D_1$ by $D_M$ in the analysis of $\delta_1/ {D_1 \choose 2}$ implies that if we can show
that $D_M \to \infty$ in probability, then
$W_1 \to w_d$ in probability, and thus $\Delta_n \to d$ in probability as $n \to \infty$.
For a random geometric graph on the torus of $\R^d$, we have from simple considerations
that $D_M \to \infty$ in probability if for all $\epsilon > 0$
$$
\lim_{n \to \infty} nr_n^{d(1-\epsilon)} = \infty~.
$$
A sufficient condition for this is that $n r_n^d \ge L(n)$
with $L(n)$ slowly varying. See Bingham, Goldie and Teugels \cite{TeBiGo87} for more on this topic.
We have $D_M \ge D_1$, and thus $D_M \to \infty$ in probability when $D_1 \to \infty$ in probability.
As $D_1$ is binomial $(n-1, V_d r_n^d)$, where $V_d$ denotes
the volume of the unit ball in $\R^d$, we have $D_1 \to \infty$ in probability when 
$n r_n^d \to \infty$. When $n r_n^d \to c > 0$ for a constant $c$, we know that
$D_M \sim {\log n / \log \log n}$ in probability by a Poissonization argument.
Thus, to show that $D_M \to \infty$, we just need to consider the case $n r_n^d \to 0$.

Fix an arbitrary large integer $t$. 
Then $D_M < t$ means that for each data point,
the $t$-th nearest neighbor is at least distance $r$ away.
So, we grid the torus with cubes of side length $\rho \defeq r_n/(2\sqrt{d})$,
which ensures that each cell in the grid can at most have $t$ data points.
As the cardinalities of the cells jointly form a multinomial random
vector, and the multinomial components are negatively associated,
we have
\begin{eqnarray*}
\PROB \{ D_M < t \} 
&\le & \PROB \{ \text{all cells have} \ \le t \ \text{data points} \} \\
&\le & ( \PROB \{ \text{Binomial}(n, \rho^d) \le t \} )^{1/\rho^d} \\
&\le & \exp \left(- {\PROB \{ \text{Binomial}(n, \rho^d) > t  \} \over \rho^d} \right) \\
&\le & \exp \left(- { {n \choose t+1} \rho^{d(t+1)} (1 - \rho^d)^{n-t-1} \over \rho^d} \right)~. 
\end{eqnarray*}
The absolute value of the exponent is of asymptotic order
$$
  { (n \rho^d )^{t+1}  \over \rho^d } = \Theta \left( n^{t+1} r_n^{dt} \right)~,
$$
and this tends to $\infty$ as $n \to \infty$ by our condition. $\square$

\medskip
\noindent
{\bf Proof of (ii).}
We rewrite the estimate as
$$
W_2 \defeq A_1 / A_2~,
$$
where
$$
A_1 = {1 \over {n \choose 3} } \sum_{1 \le i < j < k \le n} \xi_{ki} \xi_{kj} \xi_{ij}
$$
and
$$
A_2 = {1 \over {n \choose 3} } \sum_{1 \le i < j < k \le n} \xi_{ki} \xi_{kj}~.
$$
We observe that $\EXP \{ A_1 \} = w_d \pi_n^2$ and $\EXP \{ A_2 \} = \pi_n^2$,
where $\pi_n \defeq V_d r_n^d$.
The ratio of these means is $w_d$.  By bounding the variances of both 
we shall show that $A_1/\EXP \{ A_1 \} \to 1$ and $A_2/\EXP \{ A_2 \} \to 1$ in probability,
so that $A_1 / A_2 \to w_d$ in probability, as required.

We begin with 
$$
\Var \left\{ \sum_{i < j < k} \xi_{ki} \xi_{kj} \right\}
= \EXP \left\{ \left( \sum_{i < j < k} (\xi_{ki}-\pi_n) (\xi_{kj}-\pi_n) \right)^2 \right\}~.
$$
Let $s$ be the set $\{ i,j,k\}$ and let $s'$ be the set $\{i',j',k'\}$.
Observe that if $| s \cap s' | \le 1$, then $\xi_{ki}, \xi_{kj}, \xi_{k'i'}, \xi_{k'j'}$
are independent. When expanding the squared expression, we are left with contributions
coming from the cases when $| s \cap s' | \ge 2$. If the intersection is of size three,
then the two ordered triples are identical. This yields a term equal to
$$
\EXP \left\{ \sum_{i < j < k} (\xi_{ki}-\pi_n)^2 (\xi_{kj}-\pi_n)^2\right\} = {n \choose 3} \pi_n^2 (1-\pi_n)^2 \le n^3 \pi_n^2.
$$
When $| s \cap t | = 2$, the graph formed by $(k,i), (k,j), (k',i'), (k',j')$ is either a tree (in fact, a star on
four vertices) or
a 4-cycle.
Only in the latter case do we have dependence and a non-vanishing contribution.
To see this, note that, for example, if $k=k'$ and $i=i'$, then the corresponding
term equals
\[
   \EXP \left\{ (\xi_{ki}-\pi_n)^2 (\xi_{kj}-\pi_n) (\xi_{kj'}-\pi_n) \right\}= 0~.
\]
On the other hand, in the case of a 4-cycle, for example, when $i=i'$ and $j=j'$, then
the corresponding term may be bounded as follows:
\begin{eqnarray*}
  \EXP \left\{ (\xi_{ki}-\pi_n) (\xi_{kj}-\pi_n) (\xi_{k'i}-\pi_n) (\xi_{k'j}-\pi_n) \right\}
    & \le & \EXP \left\{ |\xi_{ki}-\pi_n||\xi_{kj}-\pi_n| |\xi_{k'i}-\pi_n| \right\} \\
     & =& \EXP \left\{ |\xi_{ki}-\pi_n|\right\}^3 \le \pi_n^3~.
\end{eqnarray*}
Therefore, the contribution to the variance coming from the 4-cycles is at most  $n^4 \pi_n^3$.
We conclude that
$$
\Var \left\{ \sum_{i < j < k} \xi_{ki} \xi_{kj} \right\}
\le n^3 \pi_n^2 + n^4 \pi_n^3~.
$$
As
$$
\left( \EXP \left\{ \sum_{i < j < k} \xi_{ki} \xi_{kj} \right\} \right)^2 = {n \choose 3}^2 \pi_n^4~,
$$
Chebyshev's inequality shows that $A_2/\EXP \{ A_2 \} \to 1$ in probability whenever
$$
n^3 \pi_n^2 \to \infty~.
$$

The reasoning for $A_1$ is similar. When expanding
$$
\Var \left\{ \sum_{i < j < k} \xi_{ki} \xi_{kj} \xi_{ij} \right\}
= \EXP \left\{ \left( \sum_{i < j < k} (\xi_{ki}-\pi_n) (\xi_{kj}-\pi_n) (\xi_{ij} - \pi_n) \right)^2 \right\},
$$
we once again only need to consider triples $s$ and $s'$ with $| s \cap s' | \ge 2$.
When the intersection is of size $3$, the contribution to the variance is $O(n^3 \pi_n^2)$.
When the intersection is of size $2$, after breaking up the two cycles
in the graph formed by the five edges involves, the contribution to the variance is 
easily seen to be $O(n^4 \pi_n^3)$. Arguing as for $A_2$, we conclude that
$A_1/\EXP \{ A_1 \} \to 1$ in probability whenever $n^3 \pi_n^2 \to \infty$. $\square$

\section{General densities}
\label{sec:gen}

We prove the following theorem, which implies Theorem \ref{thm2}.

\begin{theorem}
\label{thm3}
Let the density $f$ be arbitrary. Assume that $r_n = o(1)$  and $n r_n^d \to \infty$.
Then
$W_4 \to w_d$ (and thus $\Delta_n \to d$) in probability as $n \to \infty$. 
\end{theorem}

\proof
We condition on $X_1$ and $D_1$, and let $Y_1,\ldots, Y_{D_1}$ be i.i.d.\ random vectors drawn
from $f$ restricted to the ball $B(X_1,r_n)$. Set $\xi_{ij} = \IND_{ ||Y_i - Y_j || \le r_n}$.
Let $Z_1, Z_2, \ldots$ be i.i.d.\ uniform random variables on $B(X_1,r_n)$.
Then
$$
\delta_1 = \sum_{1 \le i < j \le D_1} \xi_{ij}
$$
and therefore,
$$
\EXP \left\{ W_4 \mid X_1, D_1 \right\}
= \IND_{D_1 \ge 2} \times \PROB \{ ||Y_1 - Y_2 || \le r_n |X_1 \}~.
$$
Set  $\pi_n = V_d r_n^d$ and $\mu_n(x) = \int_{B(x,r_n)} f$, $x \in \R^d$.
The density of $Y_1$ given $X_1$ is given by
$$
 \frac{f(y)}{\mu_n(X_1) } \IND_{y \in B(X_1,r_n)}~.
$$
The total variation distance $\hbox{\rm TV} (Y_1, Z_1)$ given $X_1$ is
\begin{eqnarray*}
{1 \over 2} \int_{B(X_1,r_n)} \left| { f(y) \over \mu_n(X_1) } - { 1 \over \pi_n  } \right| \, dy
&=& {1 \over 2\mu_n(X_1)} \int_{B(X_1,r_n)} \left| f(y) - { \mu_n(X_1) \over \pi_n } \right| \, dy \\
&\defeq & {\psi_n (X_1) \over 2\mu_n(X_1)}. \\
&= & {\psi_n (X_1) \over \pi_n} \times { \pi_n \over 2\mu_n(X_1)}. 
\end{eqnarray*}
The Lebesgue density theorem (see, e.g., Wheeden and Zygmund \cite{WhZy77}),
implies that for almost all $x$,
$$
\lim_{r \downarrow 0} {1 \over V_d r^d} \int_{B(x,r)} f(y) dy = f(x)~.
$$
Thus, as $r_n \downarrow 0$, $\psi_n (x)/\pi_n  \to 0$ as $n \to \infty$ for almost all $x$.
Similarly, $\mu_n(x)/\pi_n \to f(x)$ as $n \to \infty$ for almost all $x$.
We can couple $Y_1$ and $Z_1$ such that, for any $x$ with $f(x)>0$, conditional on $X_1 = x$,
$$
\PROB \{ Y_1 \not= Z_1 \} = o(1)
$$
for almost all $x$. Similarly, we can couple $Y_2$ with a uniform random vector $Z_2$.
Thus,
\begin{eqnarray*}
| \PROB \{ ||Y_1 - Y_2 || \le r_n \} - w_d |
&=& | \PROB \{ ||Y_1 - Y_2 || \le r_n \} - \PROB \{ ||Z_1 - Z_2 || \le r_n \} | \\
&\le & \PROB \{ [Y_1 \not= Z_1] \cup [Y_2 \not= Z_2]  \}  \\
&= & \int f(x)  \PROB \{ [Y_1 \not= Z_1] \cup [Y_2 \not= Z_2] |X_1 =x  \}\,dx  \\
&= & o(1)
\end{eqnarray*}
by the Lebesgue dominated convergence theorem.

From the discussion above, it helps to define the mean
$$
\nu_n (x) = \EXP \{ \xi_{12} | X_1 = x \}~. 
$$
We have
\begin{eqnarray*}
\Var \{ \delta_1 | X_1, D_1 \} 
&= & \IND_{D_1 \ge 2} \EXP \left\{ \left( \sum_{1 \le i < j \le D_1} (\xi_{ij}-\nu_n (X_1)) \right)^2 | X_1, D_1 \right\} \\
&= & \IND_{D_1 \ge 2} {D_1  \choose 2} \EXP \left\{ (\xi_{12}-\nu_n (X_1))^2 | X_1 \right\} \\
&\le & \IND_{D_1 \ge 2} {D_1  \choose 2} \nu_n (X_1)~.
\end{eqnarray*}
Finally, for arbitrary $\epsilon > 0$,
\begin{eqnarray*}
\PROB \left\{ \left| W_4 - w_d \right| > 2\epsilon \mid X_1, D_1 \right\} 
&\le & \IND_{D_1 \ge 2} \PROB \left\{ \left| W_4 - \nu_n(X_1) \right| > \epsilon \mid X_1, D_1 \right\}  \\
  & &  + \IND_{D_1 \ge 2} \PROB \left\{ \left| \nu_n(X_1) - w_d \right| > \epsilon \mid X_1 \right\}  + \PROB \{ D_1 \le 1 \} \\
&\defeq & I + II + III. 
\end{eqnarray*}
We have
$$
\EXP \{ II \} 
\le \int f(x) \PROB \left\{ \left| \nu_n(x) - w_d \right| > \epsilon \right\} \, dx
\to 0
$$
as $n \to \infty$ by the Lebesgue dominated convergence theorem, since $\nu_n(x) \to w_d$ as
$n \to \infty$ for almost all $x$.
By Chebyshev's inequality,
$$
\PROB \left\{ \left| W_4 - \nu_n(X_1) \right| > \epsilon \mid X_1, D_1 \right\}
\le {1 \over \epsilon^2} \EXP \left\{ \left( W_4 - \nu_n(X_1) \right)^2  \mid X_1, D_1 \right\} 
\le {1 \over \epsilon^2} \IND_{D_1 \ge 2} {\nu_n(X_1) \over {D_1 \choose 2} }~, 
$$
and therefore
$$
\EXP \{ I \}
\le {1 \over \epsilon^2} \int f(x) \nu_n(x) dx \times \EXP \left\{ {\IND_{D_1 \ge 2} \over {D_1 \choose 2}} \right\}
\le {1 \over \epsilon^2} \EXP \left\{ { \IND_{D_1 \ge 2} \over {D_1 \choose 2}} \right\}
$$
which is $o(1)$ if $D_1 \to \infty$ in probability. Finally, $\EXP \{ III \} \to 0$ under the
same condition on $D_1$. We conclude by noting that $D_1 \to \infty$ in probability
if $n r_n^d \to \infty$, as $D_1$ is binomial $(n-1, \mu_n(X_1))$. So, for any fixed $t$,
$$
\PROB \{ D_1 \le t \}
\le \PROB \{ (n-1) \mu_n(X_1)  \le 2t \} + \PROB \{ \hbox{\rm binomial} ((n-1), 2t/(n-1)) \le t \}
\defeq A + B.
$$
Clearly, $B \le 2/t$ by Chebyshev's inequality.
Noting that $\mu_n(x) = \pi_n (\mu_n(x)/\pi_n)$ and $\mu_n(x)/\pi_n \to f(x)$ at almost all $x$ as $n \to \infty$,
we have for arbitrary $\epsilon > 0$, 
\begin{eqnarray*}
A
&=& \PROB \{ (n-1) \mu_n(X_1)  \le 2t \}  \\
&\le & \IND_{ (n-1) \pi_n  \le 1/\epsilon } + \PROB \left\{ {\mu_n(X_1) \over \pi_n}  \le 2t\epsilon \right\}  \\
&= & \int f(x) \IND_{ {\mu_n(x) / \pi_n}  \le 2t\epsilon }  + o(1) \\
&\le & \int f(x) \IND_{ f(x)  \le 3t\epsilon }  + o(1) 
\end{eqnarray*}
which can be made as small as desired by our choice of
$\epsilon$. Hence, for any fixed $t>0$, $\PROB \{ D_1 \le t \} \le 2/t +
o(1)$, which implies that $D_1 \to \infty$ in probability.
$\square$

\medskip

Finally, we prove the following that implies Theorem \ref{thm1}.

\begin{theorem}
\label{thm4}  
Let the density $f$ have $\int f^5 < \infty$. Assume that $r_n = o(1)$  and $n^{3/2} r_n^d \to \infty$.
Then
$W_2 \to w_d$ (and thus $\Delta_n \to d$) in probability as $n \to \infty$. 
\end{theorem}

\proof
We use the notation $\mu_n(x) = \PROB \{ X_1 \in B(x,r_n) \}$ and $\nu_n (x) = \PROB \{ [X_1, X_2 \in B(x,r_n)] \cap [\|X_1 - X_2\| \le r_n] \}$.
We have
$$
\EXP \{ \xi_{12} \}  = \EXP \{ \mu_n (X_1) \}  = \int f(x) \mu_n (x) \, dx~.
$$
Let us introduce the maximal function 
$$
f^*(x) = \sup_{r>0} {\int_{B(x,r)} f(y) \, dy \over V_d r^d }~,
$$
and observe that $f \le f^*$ almost everywhere, and that $\int f^p < \infty$ for fixed
$p >1$ implies $\int (f^*)^p < \infty$ [see, e.g., Wheeden and Zygmund \cite{WhZy77}].
Thus, as $\int f \mu_n \le V_d r_n^d \int (f^*)^2$ and $\mu_n (x) / (V_d r_n^d) \to f(x)$
at almost all $x$ by the Lebesgue density theorem and $r_n \to 0$, the Lebesgue dominated
convergence theorem 
implies that  $\int f \mu_n \sim  V_d r_n^d \int f^2$ as $n \to \infty$.
In other words,
$$
\EXP \{ \xi_{12} \}  = V_d r_n^d \left(\int f^2 + o(1)\right)~.
$$
Next,
$$
\EXP \{ \xi_{12} \xi_{13} \}  \defeq M_n = \EXP \{ \mu_n^2 (X_1) \}  = \int f(x) \mu_n^2 (x) \, dx~.
$$
Extending the argument given above, we see that if $\int f^3 < \infty$, then
$$
\EXP \{ \xi_{12} \xi_{13} \}  = (V_d r_n^d)^2 \left(\int f^3 + o(1)\right)~.
$$
Using the coupling argument of the proof of Theorem \ref{thm3}, we can verify that
$$
\EXP \{ \xi_{12} \xi_{13} \xi_{23} \}  \defeq M'_n = \int f(x) \nu_n (x) \, dx = w_d (V_d r_n^d)^2 \left(\int f^3 + o(1) \right)~.
$$
We also need a general upper bound for  
$$
\EXP \left\{ \prod_{e \in E} \xi_e \right\}
$$
where $E$ is a fixed finite set of pairs of indices drawn from $\{ 1,2,\ldots,n\}$.  An example includes
$\EXP \{ \xi_{12} \xi_{13} \xi_{23} \xi_{24} \xi_{45} \}$.
Let $v(E)$ denote the size of the set of vertices involved in the definition of $E$, and 
assume that the graph defined by $E$ is connected. Since the graph is connected,
all vertices are at most at graph distance $v(E)-1$ from the node of smallest index.
Thus,
$$
\EXP \left\{ \prod_{e \in E} \xi_e \right\}
\le \left( V_d (v(E)-1)^d r_n^d \right)^{v(E)-1}  \int f(x) ( \mu'_n (x) )^{v(E)-1} \,dx
$$
where 
$$
\mu'_n (x) = { \int_{B(x, (v(E)-1) r_n)} f \over V_d (v(E)-1)^d r_n^d }  \le  f^*(x).
$$
As $f \le f^*$, we have
$$
\EXP \left\{ \cap_{e \in E} \xi_e \right\}
\le O\left(r_n^{d(v(E)-1)}\right)  \int (f^*)^{v(E)}~.
$$
Armed with this, we have
$$
\EXP \left\{ \sum_{1 \le i < j < k \le n} \xi_{ki} \xi_{kj} \right\} = {n \choose 3} \int f(x) \mu_n^2 (x) \, dx = { n \choose 3} (V_d r_n^d)^2 \left(\int f^3 + o(1) \right) \to \infty.
$$
Recalling that $M_n = \int f(x) \mu_n^2 (x) \, dx$, we have
$$
\Var \left\{ \sum_{1 \le i < j < k \le n} \xi_{ki} \xi_{kj} \right\} 
= \EXP \left\{ \left( \sum_{1 \le i < j < k \le n} (\xi_{ki} \xi_{kj} - M_n) \right)^2 \right\} 
= A_0 + A_1 + A_2~,
$$
where 
$$
A_0 
= \EXP \left\{ \sum_{1 \le i < j < k \le n} (\xi_{ki} \xi_{kj} - M_n)^2  \right\} 
= {n \choose 3} (M_n - M_n^2)
\le n^3 V_d^2 r_n^{2d} \int (f^*)^3~, 
$$
\begin{eqnarray*}
A_1 
&=& \EXP \left\{ \sum_{1 \le i < j < k \le n} \sum_{1 \le i' < j' < k' \le n} \IND_{|\{i,j,k,i',j',k'\}|=5} (\xi_{ki} \xi_{kj} - M_n)(\xi_{k'i'} \xi_{k'j'} - M_n )  \right\} \\
&=& \EXP \left\{ \sum_{1 \le i < j < k \le n} \sum_{1 \le i' < j' < k' \le n} \IND_{|\{i,j,k,i',j',k'\}|=5} (\xi_{ki} \xi_{kj}\xi_{k'i'} \xi_{k'j'} - M_n^2 )  \right\} \\
&\le & \EXP \left\{ \sum_{1 \le i < j < k \le n} \sum_{1 \le i' < j' < k' \le n} \IND_{|\{i,j,k,i',j',k'\}|=5} \xi_{ki} \xi_{kj}\xi_{k'i'} \xi_{k'j'} \right\} \\
&\le & O(n^5) \times O(r_n^{4d}) \times \int (f^*)^5~, 
\end{eqnarray*}
and 
\begin{eqnarray*}
A_2 
&=& \EXP \left\{ \sum_{1 \le i < j < k \le n} \sum_{1 \le i' < j' < k' \le n} \IND_{|\{i,j,k,i',j',k'\}|=4} (\xi_{ki} \xi_{kj} - M_n)(\xi_{k'i'} \xi_{k'j'} - M_n )  \right\} \\
&\le & \EXP \left\{ \sum_{1 \le i < j < k \le n} \sum_{1 \le i' < j' < k' \le n} \IND_{|\{i,j,k,i',j',k'\}|=4} \xi_{ki} \xi_{kj}\xi_{k'i'} \xi_{k'j'} \right\} \\
&\le & O(n^4) \times O(r_n^{3d}) \times \int (f^*)^4~. 
\end{eqnarray*}
By Chebyshev's inequality, we see that 
$$
{\left\{ \sum_{1 \le i < j < k \le n} \xi_{ki} \xi_{kj} \right\} 
\over
{ n \choose 3} (V_d r_n^d)^2 \int f^3} 
\to 1
$$
in probability if $A_0 + A_1 + A_2 = o( n^6 r_n^{4d} )$, which is easily verified.

Finally, we will show that
$$
{\left\{ \sum_{1 \le i < j < k \le n} \xi_{ki} \xi_{kj} \xi_{ij} \right\} 
\over
{ n \choose 3} (V_d r_n^d)^2 \int f^3} 
\to w_d
$$
in probability, so that $W_2 \to w_d$ in probability, as required.
To see this, we note that the above ratio has expected value tending to one,
while its variance tends to zero.  The variance bound mimics the bound obtained for the variance of 
$\sum_{1 \le i < j < k \le n} \xi_{ki} \xi_{kj}$. The troublesome terms involve upper bounds for
$\EXP \{ \xi_{ki} \xi_{kj} \xi_{ij} \xi_{k'i'} \xi_{k'j'} \xi_{i'j'} \}$ when $|\{ i,j,k,i',j',k'\}| \in \{4,5\}$.
But by bounding $\xi_{ij}$ and $\xi_{i'j'}$ by one, we have an expression similar to
that dealt with above, and thus, the variance tends to zero. $\square$

\medskip

We suspect that $\int f^3 < \infty$ suffices in Theorem \ref{thm4}, but this would require
a substantially longer proof. In any case, the restriction $\int f^5 < \infty$ would imply,
for example, that for the univariate beta $(a,b)$ density, we need to have $\min(a,b) > 4/5$.
Nevertheless the theorem still covers most densities, including some that are 
nowhere continuous.

\section*{Appendix: proof of Lemma \ref{lem:beta}}

We first show the following identity
\begin{equation}
\label{eq1}
w_d 
= \PROB \left\{ \beta \left( {d+1 \over 2}, {d+1 \over 2} \right) \le {1 \over 4} \right\}
+ \PROB \left\{ \beta \left( {1 \over 2}, {d+1 \over 2} \right) \ge {1 \over 4} \right\}~.   
\end{equation}
We recall the formula for the volume of $B(0,1)$ in $\R^d$:
$$
V_d \defeq {\pi^{d/2} \over \Gamma  \left( {d+2 \over 2} \right)}~.
$$
Let $X$ and $Y$ be defined as above.
It is well-known that $R \defeq \| X \|$ is distributed as $U^{1/d}$, where $U$ is uniform on $[0,1]$:
it has density $d x^{d-1}$ on $[0,1]$.
Without loss of generality, we can assume that $X= ( R, 0,0,\ldots,0)$.
Then $\|X - Y\| \le 1$ if $Y \in A \defeq B(0,1) \cap B(X,1)$.  $A$ is a loon-shaped region
formed by two spherical caps of the same size. Call one of the two spherical caps $S$.
Let $\lambda (\cdot)$ denote the volume of a set, and recall that $V_d = \lambda (B(0,1))$. We have
$$
\PROB \{ Y \in A \} = { 2 \EXP \{ \lambda (S) \}  \over V_d }~,
$$
where the volume of the spherical cap is a function of $R$.
Standard spatial integration yields
$$
\lambda (S) = \int_{R/2}^1 V_{d-1} (1-y^2)^{d-1 \over 2} \, dy~.
$$
Thus,
\begin{eqnarray*}
  \EXP \{ \lambda (S) \} 
&= &\int_0^1 d r^{d-1} \int_{r/2}^1 V_{d-1} (1-y^2)^{d-1 \over 2} \, dy \, dr \\
&= & V_{d-1} \int_0^{1/2} (1-y^2)^{d-1 \over 2} \int_0^{2y} d r^{d-1} \, dr \, dy
       + V_{d-1} \int_{1/2}^1 (1-y^2)^{d-1 \over 2} \, dy \\
&= & V_{d-1} \int_0^{1/2} (1-y^2)^{d-1 \over 2} (2y)^d \, dy
       + V_{d-1} \int_{1/2}^1 (1-y^2)^{d-1 \over 2} \, dy \\
&= & 2^{d-1} V_{d-1} \int_0^{1/4} (y(1-y)^{d-1 \over 2} \, dy
       + {V_{d-1} \over 2} \int_{1/4}^1 (1-y)^{d-1 \over 2} y^{-1/2}\, dy \\
  &= & I + II~.
\end{eqnarray*}    
Now,
$$
I = \alpha \PROB \left\{ \beta \left( {d+1 \over 2} , {d+1 \over 2} \right) \le {1 \over 4 } \right\}~,
$$
where
$$
\alpha = 2^{d-1} V_{d-1} { \Gamma^2 \left( {d+1 \over 2} \right) \over \Gamma (d+1) }~.
$$
Furthermore,
$$
II
= \alpha' \PROB \left\{ \beta \left( {1 \over 2} , {d+1 \over 2} \right) \ge {1 \over 4 } \right\}~,
$$
where
$$
\alpha' = {V_{d-1} \over 2} { \Gamma \left( {1 \over 2} \right) \Gamma \left( {d+1 \over 2} \right) \over \Gamma \left( {d+2 \over 2} \right) }~.
$$
Combining all of the above, we obtain
$$
\PROB \{ Y \in A \} 
= {2 \alpha \over V_d} \PROB \left\{ \beta \left( {d+1 \over 2} , {d+1 \over 2} \right) \le {1 \over 4 } \right\}
  + {2 \alpha' \over V_d} \PROB \left\{ \beta \left( {1 \over 2} , {d+1 \over 2} \right) \ge {1 \over 4 } \right\}~.
$$
We verify that $\alpha = \alpha' = V_d /2$, to conclude \eqref{eq1}.
The formal verification is as follows:
\begin{eqnarray*}
{2 \alpha \over V_d}
&=& {2^d V_{d-1} \over V_d} { \Gamma^2 \left( {d+1 \over 2} \right) \over \Gamma (d+1) } 
= {2^d \Gamma \left( {d+2 \over 2} \right)  \over \sqrt{\pi} \Gamma \left( {d+1 \over 2} \right) } { \Gamma^2 \left( {d+1 \over 2} \right) \over \Gamma (d+1) } \\
&=& {2^d \Gamma \left( {d+2 \over 2} \right)  \Gamma \left( {d+1 \over 2} \right) \over \Gamma \left( {1 \over 2} \right)  \Gamma (d+1) } 
= 1 
\end{eqnarray*}
by the duplication formula for the gamma function (see, e.g., Whittaker and Watson, \cite[p.240]{WhWa27}).
This is also immediate by induction on $d$.
Next,
$$
{2 \alpha' \over V_d} 
= {V_{d-1} \over V_d} { \Gamma \left( {1 \over 2} \right) \Gamma \left( {d+1 \over 2} \right) \over \Gamma \left( {d+2 \over 2} \right) } 
= {\Gamma \left( {d+2 \over 2} \right) \over \sqrt{\pi} \Gamma \left( {d+1 \over 2} \right) } { \Gamma \left( {1 \over 2} \right) \Gamma \left( {d+1 \over 2} \right) \over \Gamma \left( {d+2 \over 2} \right) } 
= 1~. 
$$
This proves \eqref{eq1}. Next, we show that
\begin{equation}
\label{eq2}
\PROB \left\{ \beta \left( {d+1 \over 2}, {d+1 \over 2} \right) \le {1 \over 4} \right\}
= {1 \over 2} \PROB \left\{ \beta \left( {1 \over 2}, {d+1 \over 2} \right) \ge {1 \over 4} \right\}~.   
\end{equation}
That would complete the beta representation in Lemma \ref{lem:beta}.
Let $B = \beta \left( {d+1 \over 2}, {d+1 \over 2} \right)$. 
Observe that
$$
\PROB \{ B \le 1/4 \} 
= {1 \over 2} \left( \PROB \{ B \le 1/4 \} + \PROB \{ B \ge 3/4 \} \right)
= {1 \over 2} \PROB \{ |2B- 1| \ge 1/2 \}~.
$$
Now, $|2B-1|$ has a density proportional to $(1-x^2)^{d-1 \over 2}$ on $[0,1]$,
and $(2B-1)^2$ is beta $(1/2, (d+1)/2)$. Thus,
$$
\PROB \{ B \le 1/4 \} = {1 \over 2} \PROB \left\{ \beta \left( {1 \over 2}, {d+1 \over 2} \right) \ge {1 \over 4} \right\}.
$$
The monotonicity claim follows easily.  
Finally, $w_d \to 0$ since 
$\beta (1/2, d ) \to 0$ in probability as $d \to \infty$. $\square$
\medskip

\medskip
\noindent
{\bf Proof of $w_d - w_{d+1} \ge d^{-(d+o(d))/2}$.}

\noindent
Observe that
$$
\beta \left( {1 \over 2}, {d-1 \over 2} \right)
\inlaw
{ G(1) \over \sum_{i=1}^d G(i) },
$$
where $G(1), G(2), \ldots$ are i.i.d.\ gamma $(1/2)$ random variables.
Thus, with this coupling,
$$
\PROB \left\{ \beta \left( {1 \over 2}, {d \over 2} \right)  \ge {1 \over 4} \right\}
=
\PROB \left\{ \beta \left( {1 \over 2}, {d-1 \over 2} \right) \ge {1 \over 4}  \right\}
- \PROB \left\{ \beta \left( {1 \over 2}, {d-1 \over 2} \right) \ge {1 \over 4} > \beta \left( {1 \over 2}, {d \over 2} \right)  \right\}~.
$$
The last summand reduces to
\begin{eqnarray*}
\PROB \left\{ \sum_{i=2}^d G(i) \le 3 G(1) < \sum_{i=2}^{d+1} G(i) \right\}
&=& \PROB \left\{ 3G(1) - G(d+1) < \sum_{i=2}^d G(i) \le 3 G(1) \right\} \\
&\ge & \PROB \{ G(d+1) \ge 6, G(1) \in [1,2] \} \PROB \left\{ \sum_{i=2}^d G(i) \le 3  \right\} \\
&\defeq & \rho \PROB \left\{ \sum_{i=2}^d G(i) \le 3  \right\}~. 
\end{eqnarray*}
As $\sum_{i=2}^d G(i)$ is gamma $((d-1)/2)$, we see that
$$
w_{d-2}-w_{d-1} 
\ge \rho \int_0^3 { x^{d-3 \over 2} e^{-x} \over \Gamma \left( {d-1 \over 2} \right) } \, dx 
\ge { \rho \over e^3 }  { 3^{d-1 \over 2} \over \Gamma \left( {d+1 \over 2} \right) }  
= d^{-{d \over 2} +o(d)}~.  \square
$$

\section*{Acknowledgements}
The authors thank Jakob Reznikov for his assistance.


\begin{thebibliography}{29}
\providecommand{\natexlab}[1]{#1}
\providecommand{\url}[1]{\texttt{#1}}
\expandafter\ifx\csname urlstyle\endcsname\relax
  \providecommand{\doi}[1]{doi: #1}\else
  \providecommand{\doi}{doi: \begingroup \urlstyle{rm}\Url}\fi

\bibitem[Aharonyan and Khalatyan(2020)]{AhKh20}
N.G. Aharonyan and V.~Khalatyan.
\newblock Distribution of the distance between two random points in a body
  from.
\newblock \emph{Journal of Contemporary Mathematical Analysis (Armenian Academy
  of Sciences)}, 55\penalty0 (6):\penalty0 329--334, 2020.

\bibitem[Appel and Russo(2002)]{ApRu02}
Martin~J.B. Appel and Ralph~P. Russo.
\newblock The connectivity of a graph on uniform points on [0, 1] d.
\newblock \emph{Statistics \& Probability Letters}, 60\penalty0 (4):\penalty0
  351--357, 2002.

\bibitem[Araya~Valdivia and De~Castro(2019)]{ArDe19}
Ernesto Araya~Valdivia and Yohann De~Castro.
\newblock Latent distance estimation for random geometric graphs.
\newblock \emph{Advances in Neural Information Processing Systems}, 32, 2019.

\bibitem[Balister et~al.(2008)Balister, Bollob{\'a}s, and Sarkar]{BaBoSa08}
Paul Balister, B{\'e}la Bollob{\'a}s, and Amites Sarkar.
\newblock Percolation, connectivity, coverage and colouring of random geometric
  graphs.
\newblock \emph{Handbook of Large-Scale Random Networks}, pages 117--142, 2008.

\bibitem[Balister et~al.(2009)Balister, Bollob{\'a}s, Sarkar, and
  Walters]{BaBoSaWa09}
Paul Balister, B{\'e}la Bollob{\'a}s, Amites Sarkar, and Mark Walters.
\newblock Highly connected random geometric graphs.
\newblock \emph{Discrete Applied Mathematics}, 157\penalty0 (2):\penalty0
  309--320, 2009.

\bibitem[Bubeck et~al.(2016)Bubeck, Ding, Eldan, and R{\'a}cz]{BuDiElRa16}
S{\'e}bastien Bubeck, Jian Ding, Ronen Eldan, and Mikl{\'o}s~Z R{\'a}cz.
\newblock Testing for high-dimensional geometry in random graphs.
\newblock \emph{Random Structures \& Algorithms}, 49\penalty0 (3):\penalty0
  503--532, 2016.

\bibitem[Casse(2023)]{Cas23}
J{\'e}r{\^o}me Casse.
\newblock Siblings in d-dimensional nearest neighbour trees.
\newblock \emph{arXiv preprint arXiv:2302.10795}, 2023.

\bibitem[Cooper and Frieze(2011)]{CoFr11}
Colin Cooper and Alan Frieze.
\newblock The cover time of random geometric graphs.
\newblock \emph{Random Structures \& Algorithms}, 38\penalty0 (3):\penalty0
  324--349, 2011.

\bibitem[Facco et~al.(2017)Facco, d’Errico, Rodriguez, and Laio]{FaErRoLa17}
Elena Facco, Maria d’Errico, Alex Rodriguez, and Alessandro Laio.
\newblock Estimating the intrinsic dimension of datasets by a minimal
  neighborhood information.
\newblock \emph{Scientific Reports}, 7\penalty0 (1):\penalty0 12140, 2017.

\bibitem[Gilbert(1959)]{Gil59}
Edgar~N. Gilbert.
\newblock Random graphs.
\newblock \emph{The Annals of Mathematical Statistics}, 30\penalty0
  (4):\penalty0 1141--1144, 1959.

\bibitem[Gilbert(1961)]{Gil61}
Edgar~N. Gilbert.
\newblock Random plane networks.
\newblock \emph{Journal of SIAM}, 9\penalty0 (4):\penalty0 533--543, 1961.

\bibitem[Gilbert(1965)]{Gil65}
E.N. Gilbert.
\newblock The probability of covering a sphere with n circular caps.
\newblock \emph{Biometrika}, 52\penalty0 (3/4):\penalty0 323--330, 1965.

\bibitem[Granata and Carnevale(2016)]{GrCa16}
Daniele Granata and Vincenzo Carnevale.
\newblock Accurate estimation of the intrinsic dimension using graph distances:
  Unraveling the geometric complexity of datasets.
\newblock \emph{Scientific Reports}, 6\penalty0 (1):\penalty0 31377, 2016.

\bibitem[Hall(1988)]{Hal88}
Peter Hall.
\newblock \emph{Introduction to the Theory of Coverage Processes}.
\newblock Wiley, New York, 1988.

\bibitem[Janson(1986)]{Jan86}
Svante Janson.
\newblock Random coverings in several dimensions.
\newblock \emph{Acta Mathematica}, 156:\penalty0 83– 118, 1986.

\bibitem[Kang and M{\"u}ller(2011)]{KaMu11}
Ross~J. Kang and Tobias M{\"u}ller.
\newblock Sphere and dot product representations of graphs.
\newblock In \emph{Proceedings of the Twenty-Seventh Annual Symposium on
  Computational Geometry}, pages 308--314, 2011.

\bibitem[Kleinberg(2007)]{Kle07}
Robert Kleinberg.
\newblock Geographic routing using hyperbolic space.
\newblock In \emph{IEEE INFOCOM 2007-26th IEEE International Conference on
  Computer Communications}, pages 1902--1909, 2007.

\bibitem[Lichev and Mitsche(2021)]{LiMi21}
Lyuben Lichev and Dieter Mitsche.
\newblock New results for the random nearest neighbor tree.
\newblock \emph{arXiv preprint arXiv:2108.13014}, 2021.

\bibitem[McDiarmid and M{\"u}ller(2011)]{McMu11}
Colin McDiarmid and Tobias M{\"u}ller.
\newblock On the chromatic number of random geometric graphs.
\newblock \emph{Combinatorica}, 31\penalty0 (4):\penalty0 423--488, 2011.

\bibitem[Penrose(2003)]{Pen03}
M.~Penrose.
\newblock \emph{Random Geometric Graphs}, volume~5 of \emph{Oxford Studies in
  Probability}.
\newblock Oxford University Press, Oxford, 2003.

\bibitem[Penrose(1999)]{Pen99}
Mathew~D Penrose.
\newblock A strong law for the longest edge of the minimal spanning tree.
\newblock \emph{The Annals of Probability}, 27\penalty0 (1):\penalty0 246--260,
  1999.

\bibitem[Reiterman et~al.(1989)Reiterman, R{\"o}dl, and
  {\v{S}}i{\v{n}}ajov{\'a}]{ReRoSi89}
J.~Reiterman, V.~R{\"o}dl, and E.~{\v{S}}i{\v{n}}ajov{\'a}.
\newblock Geometrical embeddings of graphs.
\newblock \emph{Discrete Mathematics}, 74\penalty0 (3):\penalty0 291--319,
  1989.

\bibitem[Shavitt and Tankel(2004)]{ShTa04}
Yuval Shavitt and Tomer Tankel.
\newblock Big-bang simulation for embedding network distances in euclidean
  space.
\newblock \emph{IEEE/ACM Transactions on Networking}, 12\penalty0 (6):\penalty0
  993--1006, 2004.

\bibitem[Tenenbaum et~al.(2000)Tenenbaum, Silva, and Langford]{TeSiLa00}
Joshua~B. Tenenbaum, Vin~de Silva, and John~C. Langford.
\newblock A global geometric framework for nonlinear dimensionality reduction.
\newblock \emph{Science}, 290\penalty0 (5500):\penalty0 2319--2323, 2000.

\bibitem[Teugels et~al.(1987)Teugels, Bingham, and Goldie]{TeBiGo87}
J.L. Teugels, N.H. Bingham, and C.M. Goldie.
\newblock \emph{Regular Variations}.
\newblock Cambridge University Press, 1987.

\bibitem[Verbeek and Suri(2014)]{VeSu14}
Kevin Verbeek and Subhash Suri.
\newblock Metric embedding, hyperbolic space, and social networks.
\newblock In \emph{Proceedings of the Thirtieth Annual Symposium on
  Computational Geometry}, pages 501--510, 2014.

\bibitem[Vershynin(2018)]{Ver18}
Roman Vershynin.
\newblock \emph{High-Dimensional Probability: An Introduction with Applications
  in Data Science}, volume~47.
\newblock Cambridge University Press, 2018.

\bibitem[Wheeden and Zygmund(1977)]{WhZy77}
A.~Wheeden and R.L. Zygmund.
\newblock \emph{Measure and Integral}.
\newblock Marcel Dekker, New York, 1977.

\bibitem[Whittaker and Watson(1927)]{WhWa27}
E.T. Whittaker and G.N. Watson.
\newblock \emph{A Course of Modern Analysis}.
\newblock Cambridge University Press, 1927.

\end{thebibliography}

\end{document}